\documentclass[gpscopy,onehalfspacing,11pt]{ubcdiss}

\usepackage{times,mathptmx,courier}
\usepackage[scaled=.92]{helvet}
\usepackage[compact]{titlesec}

\titleformat{\chapter}{\normalfont\huge\bf}{\thechapter.}{20pt}{\huge\bf}

\usepackage{pifont}

\usepackage{checkend}	
\usepackage{graphicx}	

\usepackage{booktabs}

\usepackage{listings}
\usepackage[printonlyused,nohyperlinks]{acronym}

\usepackage{color}
\definecolor{greytext}{gray}{0.5}

\usepackage{comment}

\usepackage[numbers,sort&compress]{natbib}
\usepackage[a4paper, total={6in, 8in}]{geometry}
\usepackage{multirow}
\usepackage{graphicx}

\titleformat*{\section}{\singlespacing\raggedright\bfseries\Large}
\titleformat*{\subsection}{\singlespacing\raggedright\bfseries\large}
\titleformat*{\subsubsection}{\singlespacing\raggedright\bfseries}
\titleformat*{\paragraph}{\singlespacing\raggedright\itshape}

\usepackage[format=hang,indention=-1cm,labelfont={bf},margin=1em]{caption}

\usepackage{url}
\ifgpscopy
  \usepackage[bookmarks,bookmarksnumbered,%
     pagebackref,linktocpage,%
     colorlinks=true,%
     linkcolor=black,%
     urlcolor=blue,%
     citecolor=black%
     ]{hyperref}
\else
  \usepackage[bookmarks,bookmarksnumbered,%
    pagebackref,linktocpage,%
    allbordercolors={0.8 0.8 0.8},%
    ]{hyperref}
\fi
\newcommand{\nocitations}{\relax}
\renewcommand*{\backrefalt}[4]{%
\textcolor{greytext}{\ifcase #1%
\nocitations%
\or
\(\rightarrow\) page #2%
\else
\(\rightarrow\) pages #2%
\fi}}

\DeclareUrlCommand\DOI{}
\newcommand{\doi}[1]{\href{http://dx.doi.org/#1}{\DOI{doi:#1}}}



\title{Reducing Impacts of System Heterogeneity in Federated Learning using Weight Update Magnitudes}

\author{Irene Wang}
\advisor{Dr. Prashant Nair}
\coAdvisor{Dr. Divya Mahajan (Microsoft Research)}

\degreetitle{Bachelor of Applied Science}

\institution{The University of British Columbia}
\campus{Vancouver}

\faculty{The Faculty Applied Science}
\department{Computer Engineering}
\course{CPEN499}
\submissionmonth{August}
\submissionyear{2022}

\hypersetup{
  pdftitle={IW_CPEN499},
  pdfauthor={Irene Wang}
}

\begin{document}


 \maketitle
\chapter{Abstract}
The widespread adoption of handheld devices have fueled rapid growth in new applications.
Several of these new applications employ machine learning models to train on user data that is typically private and sensitive.
Federated Learning enables machine learning models to train locally on each handheld device while only synchronizing their neuron updates with a server.
While this enables user privacy, technology scaling and software advancements have resulted in handheld devices with varying performance capabilities.
This results in the training time of federated learning tasks to be dictated by a few low-performance straggler devices, essentially becoming a bottleneck to the entire training process.

%
%
%
In this work, we aim to mitigate the performance bottleneck of federated learning by dynamically forming sub-models for stragglers based on their performance and accuracy feedback. 
To this end, we offer the \emph{Invariant Dropout}, a dynamic technique that forms a sub-model based on the neuron update threshold. 
Invariant Dropout uses neuron updates from the non-straggler clients to develop a tailored sub-models for each straggler during each training iteration.
%
All corresponding weights which have a magnitude less than the threshold are dropped for the iteration.
%
%
We evaluate Invariant Dropout using five real-world mobile clients.
Our evaluations show that Invariant Dropout obtains a maximum accuracy gain of 1.4\% points over state-of-the-art Ordered Dropout while mitigating performance bottlenecks of stragglers.
\cleardoublepage

\tableofcontents
\cleardoublepage	

\listoftables
\cleardoublepage	

\listoffigures
\cleardoublepage	



\textspacing		

\acresetall	
\mainmatter

\chapter{Introduction}
\label{ch:Introduction}

In the past decade, relentless technology scaling and micro-architectural innovations have resulted in computationally powerful handheld devices~\cite{Li2020FederatedLC,flee}.
Handheld devices execute applications that process data to offer a personalized experience to the users~\cite{shankar2009mobile,shankar2010mobile}. 
These applications typically use Machine Learning (ML) models to train on data that is private to each individuals' handheld device.  
To maintain user privacy, rather than transferring individual data into a server that hosts the ML model, models are typically trained locally on each handheld device -- also called a client~\cite{FedAvg}.
%
%
This process is called Federated Learning (FL) and presently, it is employed by several industry players such as Meta, Google, etc~\cite{FedBuff,FedAvg, Gboard}.

There is high variation in the computational power of handheld clients.
This poses a key challenge for FL. 
For instance, our thesis observes that clients that are even a few years apart tend to offer dramatically different computational performance.
Unfortunately, clients that offer low computational performance dictate the training time and accuracy of FL, and are labelled as stragglers.
This thesis aims to mitigate the computational bottlenecks of stragglers while providing high accuracy.

Depending on the accuracy and performance goals, FL can be implemented using two approaches. 
First, FL can employ an asynchronous aggregation protocol~\cite{fedAT,ASO-Fed,FedAsync}.
This approach enables clients to update the server model asynchronously.
While this mitigates the effects of stragglers, it has the problem that there could be instances when the model is stale~\cite{Barkai2020Gap-Aware}.
Therefore, asynchronous aggregation protocol-based FL tend to display slower convergence and lower accuracy~\cite{fedAT,FedAdapt,tamingMomentum}.
%
Second, FL can employ a synchronous aggregation protocol such as FedAvg~\cite{FedAvg}. 
In this approach, the server waits for all clients to share their updates in each training iteration and thereby maintains an updated server model. 
This prevents instances of stale models and improves accuracy at the expense of stragglers dictating the training time ~\cite{fjord}.
As ML model accuracy is key, this thesis focuses on synchronous aggregation protocol-based FL.

Prior works advocate sending a smaller subset of the server model to the stragglers ~\cite{FedDrop,fjord}. This helps avoid training bias from uneven client sampling and allows stragglers to contribute towards training the model~\cite{fjord, advancesinFed}. 
As the stragglers are now training on a subset of the server model, they showcase a reduced training time.
This helps reduce the computational effect of stragglers on the overall training time -- albeit with a loss of accuracy as only a portion of the server model is being trained on stragglers.

To this end, techniques such as Federated Dropout~\cite{FedDrop} randomly drop parts of the server model and send a subset of the model to the stragglers.
However, as the neurons in the server model are randomly dropped, Federated Dropout incurs an accuracy drop.
A state-of-the-art work, called Ordered Dropout from FjORD ~\cite{fjord}, reduces this accuracy loss by regulating the neurons to be dropped.
Ordered Dropout drops either the right or left portions of the server model, thereby ensuring neurons that are not dropped maintain their connections.
Thus, the subset of neurons that are dropped from the server model dictates the overall accuracy of the learning problem.

\begin{figure}[t!]
    \centering
    \includegraphics[width=0.8\columnwidth]{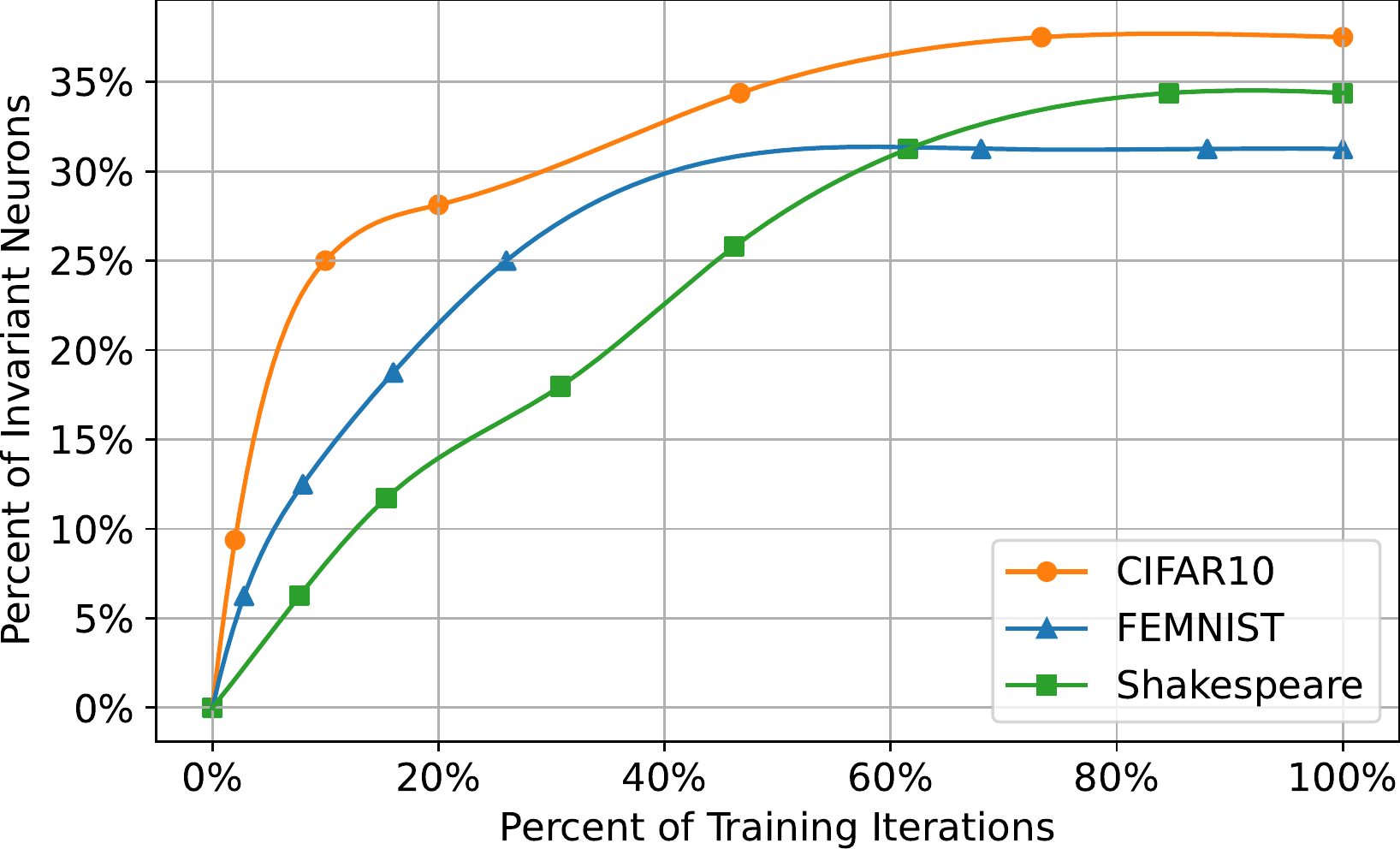}
   \caption[Percentage of `invariant' neurons vs. Iterations]{The percentage of `invariant' neurons as the number of training iterations vary for the CIFAR10~\cite{cifar10}, FEMNIST~\cite{Leaf}, and Shakespeare~\cite{Leaf} Datasets. Overall, even after 30\% of the training iterations, 15\%-30\% of the neurons remain invariant across the three datasets. For computing the invariant neurons, we choose thresholds of  180\%, 10\%, and 500\% respectively for these three datasets.}
    \label{fig:intro}
\end{figure}

Our thesis uses the insight that some neurons in the server model are trained quickly and vary only slightly -- which we call `invariant' neurons. 
Figure~\ref{fig:intro} shows the percentages of invariant neurons as the number of training iterations increases.
%

Even after only 30\% of the training iterations are completed, 15\%-30\% of the neurons become invariant across CIFAR10~\cite{cifar10}, FEMNIST~\cite{Leaf}, and Shakespeare~\cite{Leaf} datasets.
For this example, we choose thresholds of 180\%, 10\%, and 500\% respectively for these three datasets and compute their invariant neurons.
Sending invariant neurons over to the straggler provides no utility and therefore these neurons can be dropped.
We use the magnitude of neuron updates to identify the invariant neurons.
This is challenging as it would require the server to anticipate which neurons would change while only supplying a sub-model to the straggler.
Our thesis uses the non-stragglers to overcome this challenge.

We observe that, by design, the the number of non-straggler clients tend to be larger in number as compared to straggler clients.
Additionally, non-straggler clients train on the entire model.
Thus, by design, they can provide directions on which neurons are invariant.
Over time, beside the invariant neurons, all other neurons tend to be updated across on all clients.
Using this insight, this thesis proposes Invariant Dropout, a technique that prioritizes the dropout of neurons that change below a certain threshold -- called drop-threshold.
The drop-threshold is dynamically determined based on the neuron update characteristics. 
The target sub-model size is determined by profiling client training times in the first training iteration and identifying the straggler(s). 
The drop-threshold is then varied as training iterations increase until the amount of neurons that are chosen equals the target sub-model size for the straggler(s). 

Overall, this thesis makes four key contributions: 

\begin{enumerate}
    \item We propose a tailored straggler identification method that dynamically determines the sub-model size to be executed on stragglers based on their individual performance overheads. This mitigates the performance overheads of stragglers.  
    \item We introduce Invariant Dropout. Invariant Dropout dynamically selects a subset of the server model to train on the straggler based on magnitude of neuron updates over a dynamically determined threshold.
    \item We implement Invariant Dropout, Federated Dropout, and Ordered Dropout on five mobile phones. This implementation showcases the real-world effect of stragglers on computational performance and accuracy.
    \item We show that Invariant Dropout improves training accuracy with a maximum 1.5\% points over state-of-the-art Ordered Dropout while mitigating the computational performance overheads of stragglers.
\end{enumerate}


\chapter{Background and Motivation}
\label{ch:Motivation}
We briefly describe the background on Federated Learning and provide motivation for mitigating stragglers.
\section{Global Server and Clients}
Federated Learning (FL) is typically performed using centralized global servers and distributed clients, typically handheld devices. 
In FL systems using synchronous aggregation protocols like FegAvg~\cite{fedAT}, the server maintains a central copy of the ML model called the global model. 
The clients contain private user data and the server sends the global model to each client at the beginning of each training iteration.
At the end of each iteration, the server aggregates the neuron updates from each client into the global model.
The computational performance of each client varies as they can span different technology generations.
This causes some clients to showcase low computational performance and be denoted as stragglers.
%
%

\section{Dropout Techniques}
Several studies have focused on ensuring privacy and robustness~\cite{Zhao2022FedInvBF, Bonawitz2017PracticalSA,Gong2022PreservingPI,McMahan2018LearningDP}, data heterogeneity~\cite{Liang2020ThinkLA}, and improving accuracy for FL~\cite{FedAvg,FedProx,Yan2022SeizingCL,Wang2020TacklingTO, reddi2021adaptive}. 
However, another important challenge in FL is system heterogeneity amongst client devices~\cite{Ding2022FederatedLC, Li2020FederatedLC}.
Due to rapid technological development in both hardware and software, even client devices that are just a few years apart can differ dramatically in terms of processing abilities and memory capacity. 
Prior work has shown that, simply discarding slower or lower-tier clients may introduce bias in the trained model~\cite{fjord, advancesinFed}.
Yet, including stragglers in the FL network slows down the training process, and causes more powerful devices to stall for long periods of time between training iterations. 
To this end, there are two state-of-the-art techniques.
\subsection{Random Dropout} Federated Dropout~\cite{FedDrop} has proposed sending a smaller subset of the global model to the straggler clients.
These work advocate dropping neurons in a random fashion and thus effect the accuracy of the global model.

\subsection{Ordered Dropout} Ordered Dropout from FjORD~\cite{fjord} aims to systematically drop neurons by preserving order in the sub-model. 
This is done by selecting only parts of the right or the left portions of the global model to create a sub-model.
This state-of-the-art work establishes that the selection of neurons in the sub-model plays a key role in the overall accuracy of the trained model. 
By neurons, this work refers to the filters from convolutional (\texttt{CONV}) layers, activations from fully-connected (\texttt{FC}) layers, and hidden units from LSTM~\cite{LSTM} layers.
This setup assumes there is one straggler device. 
The sub-model size for Ordered Dropout is varied from 0.5 to 1 (i.e. complete global model).
As the sub-model size reduces, across three datasets and ML models, Ordered Dropout incurs up to 4.5\% point drop in accuracy.
\section{Motivation 1: Accuracy Implications}
\begin{figure}[h]
    \centering
    \includegraphics[width=0.8\columnwidth]{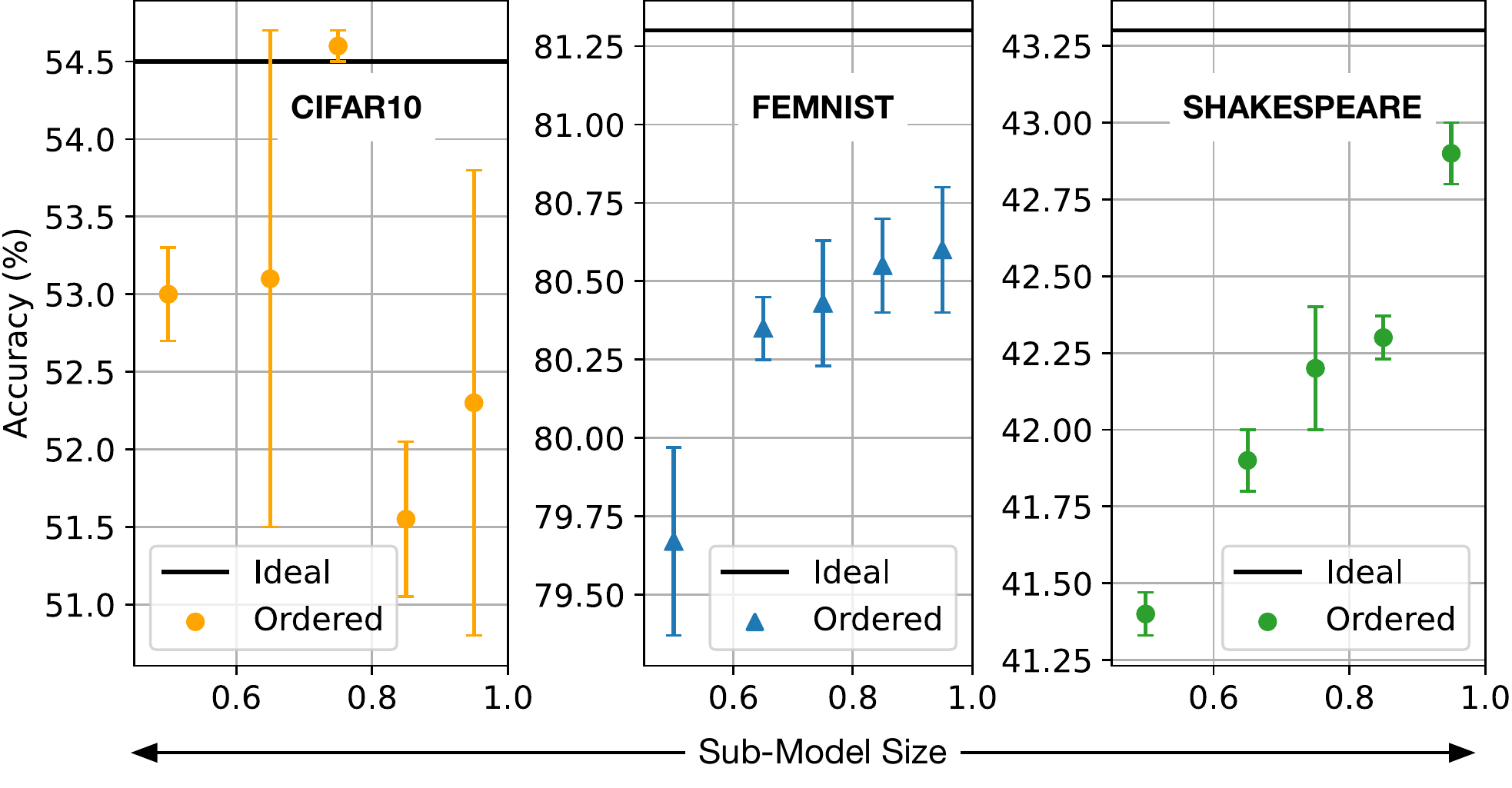}
   \caption[Ordered Dropout Accuracy Comparison with Ideal Scenario]{The relative difference in accuracy between an ideal implementation of Federated Learning as compared to Ordered Dropout. Overall, Ordered Dropout shows up to 4.5\% point drop in accuracy.}
    \label{fig:motivation1}
\end{figure}

Figure~\ref{fig:motivation1} shows the difference in testing accuracy between an `ideal' FL implementation and Ordered Dropout by using 5 mobile devices using the CIFAR10, FEMNIST, and Shakespeare datasets. 
Ideally, we would like the dropout technique to display an accuracy that is closer to the `ideal' FL implementation.

\section{Motivation 2: Performance Implications}
\begin{figure}[h!]
    \centering
    \includegraphics[width=0.8\columnwidth]{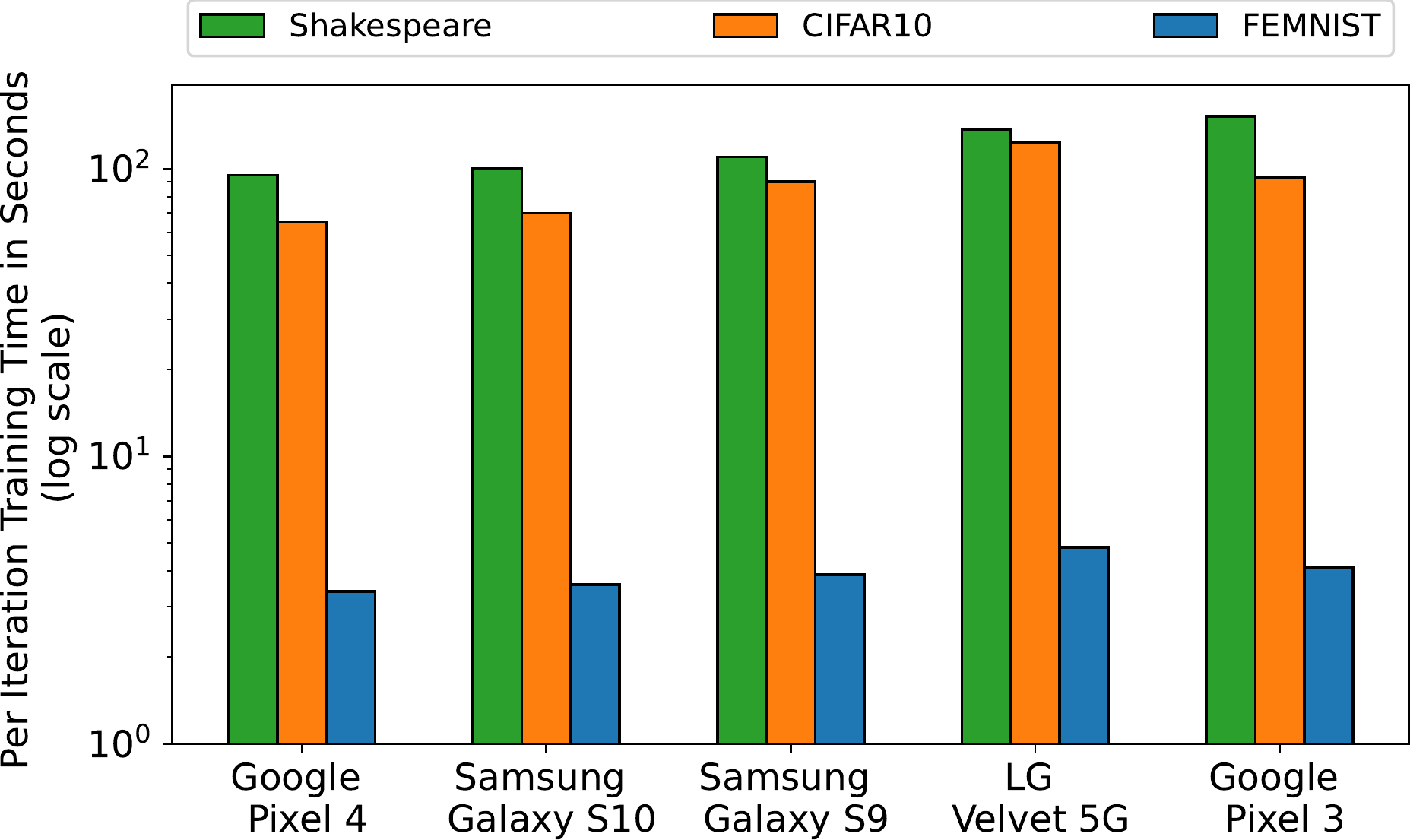}
   \caption[Per-iteration Training Time of Client Devices]{The per-iteration training time of client devices. We use five android-based mobile devices with varying performance characteristics. The choice of straggler is not fixed and it depends on the dataset and ML model. Furthermore, depending on the dataset, the per-iteration training time can vary from $<$10 seconds to $>$100 seconds.}
    \label{fig:motivation2}
\end{figure}

Figure~\ref{fig:motivation2} shows the difference in the overall training time of 5 mobile devices involved in FL without using any dropout techniques.
Depending on the dataset and ML model, a client might display less than 10 seconds or greater than 100 seconds of training time.
Furthermore, stragglers are not fixed and each client may become a straggler for any particular dataset and ML model.
For instance, while Google Pixel 3 is the straggler for the Shakespeare dataset using a two-layer LSTM classifier.
LG Velvet 5G is the straggler for the FEMNIST dataset applying a CNN model.
Ideally, we would like to dynamically identify stragglers and enable them to showcase the training time that is closer to the non-straggler devices.

\begin{figure}[h!]
    \centering
    \includegraphics[width=1\columnwidth]{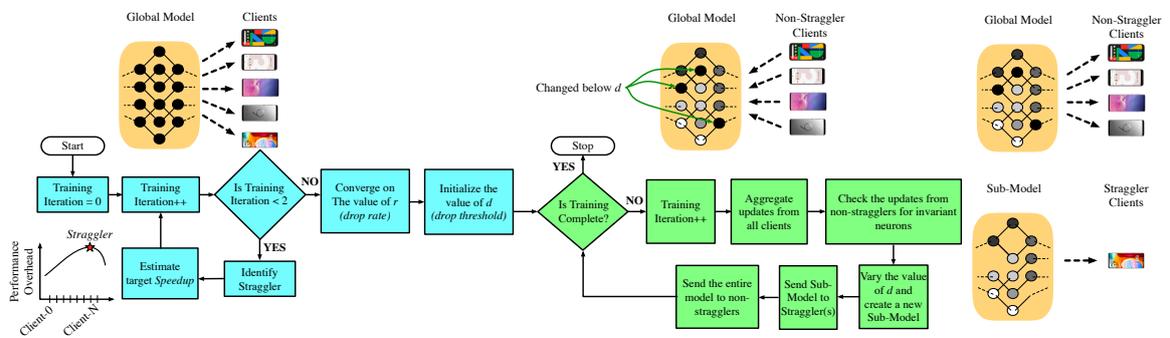}
   \caption[Invariant Dropout High-level Overview]{The high-level flow of Invariant Dropout. The non-stragglers inform the global model of the neurons that are not updated within a set threshold. Thereafter, as the training process continues, sub-models are dynamically created by dropping invariant neurons. These sub-models are sent to the straggler devices.}
    \label{fig:overview}
\end{figure}
 \vfill

\chapter{Design: Invariant Dropout}
\label{ch:Design}

This thesis introduces Invariant Dropout, a technique that lets straggler clients train with a sub-model that is dynamically generated from neurons that `vary' over time. 
Figure~\ref{fig:overview} shows the flow of Invariant Dropout.
Invariant Dropout consists of three steps.

\section{Identifying the Performance of Stragglers} Invariant Dropout initially executes the global model on every client, including stragglers.
This step occurs in the first training iteration of each dataset and ML model.
After the first iteration, the global server samples the performance delay between the slowest client's (straggler) training time ($T_{straggler}$) and the target training time ($T_{target}$).
When there is only one straggler, $T_{target}$ is set as the training time of the `next slowest client'. 
The $Speedup$ that's required for the stragglers can be computed using Equation~\ref{eqn:speedup}.

\begin{equation}
    Speedup = \frac{T_{straggler}}{T_{target}}
    \label{eqn:speedup}
\end{equation}

\section{Tuning the Performance of Stragglers}
The dropout rate, denoted as $r$, is varied based on the value of $Speedup$.
The value of $r$ is between $0$ and $1$, and it determines the sub-model size that is trained on the straggler.
$r$ is chosen such that the updated straggler training time, denoted as $T_{straggler-new}$, and can be derived from Equation~\ref{eqn:speedup} by enabling $Speedup$ to be close to 1; as shown in  Equation~\ref{eqn:speeduptarget}.

\begin{equation}
      T_{straggler-new} \approx T_{target}
    \label{eqn:speeduptarget}
\end{equation}

%

The dropout rate is kept constant for all layers in the model. 
During each training iteration, the server constructs the sub-model by dropping (1-$r$)\% of the neurons per layer.

\section{Drop Threshold}
Invariant Dropout dynamically maintains an update threshold value, below which a neuron can be classified as invariant.
We define the percent difference of a neuron's weight update.
Assume the $w_{ijc}(t)$ represents the set of weight parameters of the $i$th neuron in the $j$th layer for client $c$ after the training iteration $t$. The percent difference $d$ of the the neuron, for each client $c$, is the minimum value of $d$ as denoted by Equation~\ref{eqn:diff}.
\begin{equation}
      d \ge \frac{w_{ijc}(t) - w_{ij}(t -1)}{w_{ij}(t -1)}
    \label{eqn:diff}
\end{equation}

Neurons whose weight updates lie within $d$ as compared to the previous training iteration are chosen as potential drop candidates.
Unfortunately, this is non-trivial to implement.

\section{Dynamically Identifying Drop Thresholds}
The global server cannot determine the drop candidates by simply checking the updates from the stragglers.
This is because, the stragglers only train on a sub-model and therefore would only update parts of that sub-model.
To overcome this concern, the global server uses the insight that the non-straggler clients train on the complete global model and therefore have access to all the neurons.
The non-stragglers clients can therefore be used to identify neurons whose weight updates lie within $d$, at a per-iteration level.
Invariant Dropout prioritizes dropping neurons whose weight updates fall within $d$ for the \emph{majority} of the non-straggler clients and dynamically create a sub-model for each training iteration.
%

%
%
%

Furthermore, the initial value of $d$ is chosen such that it is average of the minimum percent update out of all neurons in the layer in the first training iterations.
We observe that the percent difference of neurons change drastically during the initial stages of training.
Thus, Invariant Dropout conservatively maintains the same update threshold in the initial training iterations.
The threshold is then incrementally increased after each training iteration until the number of neurons that fall under the threshold becomes equal to or greater than the number of neurons that need to be left out of the sub-model -- as dictated by Equation~\ref{eqn:speeduptarget}.
It should be noted that Invariant Dropout enables the drop threshold to be different for each layer of the model.
This is because it can track the magnitude of weight updates for each individual layer.
Finally, to ensure that fully trained neurons are dropped instead of critical weights, the algorithm prioritizes leaving out neurons if they repeatedly fall within the threshold in previous iterations. 

\chapter{Convergence of Invariant Dropout}
\label{ch:Proof}




%
%

Similar to the prior work~\cite{GradientSF}, although dropout can reduce the communication cost of stragglers but can increase the gradient variance.
We prove convergence of our invariant dropout technique using bounded gradient variance~\cite{GradientSF, helios}.
Consider $N$ devices participating in Federated Learning. 
The training data is represented as $\{x_{n}\}_{n=1}^{N}$, where $n$ represents a device, and each one has a loss function $\{f_{n}\}_{n=1}^{N}$.
We solve to the minimize the loss function using the following optimization:

\begin{equation}
    f(w) := \frac{1}{N} \sum_{n=1}^{N} f_n(w)
\end{equation}

\begin{equation}
    w_{t+1} = w_{t} - \eta_t(g(w_t))
\end{equation}

The invariant dropout problem can be viewed as a sparse stochastic gradient vector, where each gradient has a certain probability to be dropped. 
The gradient vector is $G = [g_1, g_2, ...,  g_m]$ where $g \in R$ with probability of staying and being transmitted across the networks as $[p_1, p_2, ...,  p_m]$.
Hence, the probability of dropout is $1 - p_i$, where $0 < i \le m$.
The sparse vector for the stragglers is represented as $G_s$.
The variance of the invariant-dropout based gradient vector can be represented as follows: 

\begin{equation}
    E\sum_{i=1}^{m}[(G_s)]^2 = \sum_{i=1}^{m} (\frac{g_i^2}{p_i})
\end{equation}

The variance of the dropout vector is a small factor deviation from the non-dropout gradient vector represented as follows: 

\begin{equation}
   \min \sum_{i=1}^{m} p_i : \sum_{i=1}^{m} (\frac{g_i^2}{p_i}) = (1 + \epsilon) \sum_{i=1}^{m}{g_i^2}
   \label{eq:variancesparisityoptimization}
\end{equation}

This aims to minimize the probability of selecting gradients, i.e. higher benefits from skipping gradients, whilst ensuring the variance being small deviation from the dense gradient vector. 

\section{Proof of convergence:}
With the invariant dropout, the weights are dropped based on their `percentage' change across iterations which essentially implies that weights corresponding to higher gradient values have a lower probability of being dropped and vice-versa.
This implies if $|g_i > g_j|$ then $p_i > p_j$.
The probability of a gradient getting dropped is proportional to the invariant dropout rate $r$.
Lets assume, in this case the top-k magnitude of gradient values are not dropped, hence if $G = [g_1, g_2, ...,  g_m]$ is assumed to be sorted, then $G = [g_1, g_2, ...,  g_k]$ has a $p = 1$, whereas the $G = [g_{k+1}, g_{k+2}, ...,  g_d]$ have a probability of $p_i = r|g_i|$.
This modifies the optimization problem in Equation~\ref{eq:variancesparisityoptimization} to the following: 

\begin{equation}
    \sum_{i=1}^{k} {g_i^2} + \sum_{i=k+1}^{m} \frac{|g_i|}{r} - (1 + \epsilon) \sum_{i=1}^{m}{g_i^2} = 0, r|g_i| \leq 1
\end{equation}

Which implies that

\begin{equation}
    r = \frac{\sum_{i=k+1}^{m} |g_i|}{(1 + \epsilon) \sum_{i=1}^{m}{g_i^2} - \sum_{i=1}^{k} {g_i^2}}
\end{equation}

As per the constraint $r|g_i| \leq 1$:

\begin{equation}
    |g_i|(\sum_{i=k+1}^{m} |g_i|) \leq (1 + \epsilon) \sum_{i=1}^{m}{g_i^2} - \sum_{i=1}^{k} {g_i^2}
\end{equation}

In this invariant dropout work we keep the gradients with highest magnitude, thus if the number of gradients. As per~\cite{GradientSF}, if $k << d$ the boundedness of the expected value from equation~\ref{eq:variancesparisityoptimization} can be represented as:

\begin{equation}
   \sum_{i=1}^{m} p_i, = \sum_{i=1}^{k} p_i + \sum_{i=k+1}^{m} p_i
\end{equation}

\begin{equation}
   \sum_{i=1}^{m} p_i = k + |g_i| \sum_{i=k+1}^{m} (\frac{\sum_{i=k+1}^{m} |g_i|}{(1 + \epsilon) \sum_{i=1}^{m}{g_i^2} - \sum_{i=1}^{k} {g_i^2}})
\end{equation}

\begin{equation}
   \sum_{i=1}^{m} p_i \leq k(1 + \epsilon)
   \label{eq:boundedness}
\end{equation}

This proves that the variance of gradient is bounded by Equation~\ref{eq:boundedness}.
\chapter{Evaluation Setup}
\label{ch:setup}
We describe our models, datasets, system configurations, evaluation metrics, and baselines.
\section{Models and Dataset}
Similar to state-of-the-art work Ordered Dropout~\cite{fjord}, we evaluate on three datasets: FEMNIST, Shakespeare from the LEAF datasets~\cite{Leaf}, and CIFAR10~\cite{cifar10}.

\subsection{The FEMNIST Dataset} For this dataset, we use a CNN with two 5x5 \texttt{CONV} layers (the first one has 16 channels, the second one has 64 channels, each of them followed with 2$\times$2 max-pooling), an \texttt{FC} dense layer with 120 units, and a final softmax output layer. 
The data is partitioned based on the writer of the character in non-IID setting. 
\subsection{The Shakespeare Dataset} For this dataset, we consider a two-layer LSTM classifier containing 128 hidden units. Model takes a sequence of one hot encoded 80 characters as input, and the output is a class label between 0 and 80. 
The data is partitioned based on each role in a play in non-IID setting. 
\subsection{The CIFAR10 Dataset} For this dataset, we use a VGG-9~\cite{VGG} architecture model with 6 3x3 \texttt{CONV} layers (first 2 have 32 channels, followed by two 64 channel layers and lastly two 128 channel layers), two \texttt{FC} dense layers with 512 and 256 units and a final softmax output layer. 
The data is randomly partitioned into equal sets, hence remaining IID in setting.

\section{System Configuration}
Table~\ref{tab:phones} provides the details on the phones used for the experiments.
We connect all our clients devices and our server over the same network.
We evaluate a total of five clients and identify one straggler per training run.
All the clients run on android mobile phones from the years 2018 to 2020.
The server runs separately on a windows machine.
\begin{table} [h!]
  \begin{center} 
    \caption{Software-Hardware Specifications of Clients}
    \label{tab:phones}
    \resizebox{1\columnwidth}{!}{
    \begin{tabular}{c|c|c|p{6cm}} 
      \textbf{Device} & \textbf{Year} & \textbf{Android OS Version} & \textbf{CPU (Cores)}\\
      \hline
      Google Pixel 4 & 2019 & 12 & 1$\times$2.84 GHz Kryo 485 + 3$\times$2.42 GHz Kryo 485 + 4$\times$1.78 GHz Kryo 485\\ \hline
      Google Pixel 3 & 2018 & 9 & 4$\times$2.5 GHz Kryo 385 Gold + 4$\times$1.6 GHz Kryo 385 Silver\\ \hline
      Samsung Galaxy S10 & 2019 & 11 & 2$\times$2.73 GHz Mongoose M4 + 2$\times$2.31 GHz Cortex-A75 + 4$\times$1.95 GHz Cortex-A55\\ \hline
      Samsung Galaxy S9 & 2018 & 10 & 4$\times$2.8 GHz Kryo 385 Gold + 4$\times$1.7 GHz Kryo 385 Silver\\ \hline
      LG Velvet 5G & 2020 & 10 & 1$\times$2.4 GHz Kryo 475 Prime + 1$\times$2.2 GHz Kryo 475 Gold + 6$\times$1.8 GHz Kryo 475 Silver\\
      \hline
    \end{tabular}
    }
  \end{center}
\end{table}


Invariant Dropout is implemented on top of the Flower (v0.18.0)~\cite{Flower} framework and TensorFlow Lite~\cite{TFLite} from TensorFlow v2.8.0~\cite{TensorFlow}. 
Each client executes its training run as an Android application. 
Models are defined using TensorFlow's Sequential API, and then converted into \texttt{.tflite} formats using TensorFlow Lite's Transfer Converter. 
Since the sub-model size for stragglers is dynamically determined at runtime, all model definitions for each possible sub-model size need to be included in our Android application as assets.
When a client receives a sub-model during training, the client will load the \texttt{.tflite} files for that specific model size. 

\section{Evaluation Metrics}
We report average performance (wall-clock training time), accuracy, and standard deviations across three execution runs for all experiments.
During each round of evaluation, all clients receive the global model, and report the evaluation accuracy and loss on its local data to the server.
The server calculates the overall distributed accuracy and loss by performing a weighted average based on the number of testing examples for each client.  
To reduce the effects of over-fitting, the overall accuracy for each run is determined by the accuracy of iteration with the lowest evaluation loss. 

\section{Baselines}
We compare Invariant Dropout with the following two baselines: 1) Random federated dropout~\cite{FedDrop} and 2) Ordered dropout from Fjord~\cite{fjord}.

\chapter{Results and Analysis}
\label{ch:Results}
\begin{table*}[b!]
\begin{center}
\caption[Accuracy Comparison of Dropout Techniques]{\centering Accuracy Comparison of Random Dropout, Ordered Dropout, and Invariant Dropout. The text in \textbf{\color{red}red} indicates instances when Invariant Dropout showcases the highest accuracy. ($\mu$ $=$ mean, $\sigma$ $=$ standard deviation, and $r$ $=$ sub-model as a fraction of the global model).}
\label{tab:accuracy}
\resizebox{1\columnwidth}{!}{
\begin{tabular}{|l|l|ll|ll|ll|ll|ll|}
\hline
\multirow{2}{*}{Dataset} & \multirow{2}{*}{Dropout Method} & \multicolumn{2}{l|}{$r$ $=$ 0.95}         & \multicolumn{2}{l|}{$r$ $=$ 0.85}         & \multicolumn{2}{l|}{$r$ $=$ 0.75}         & \multicolumn{2}{l|}{$r$ $=$ 0.65}         & \multicolumn{2}{l|}{$r$ $=$ 0.5}          \\ \cline{3-12} 
                                    &                                 & \multicolumn{1}{l|}{Accuracy ($\mu$)} & $\sigma$  & \multicolumn{1}{l|}{Accuracy ($\mu$)} & $\sigma$  & \multicolumn{1}{l|}{Accuracy ($\mu$)} & $\sigma$  & \multicolumn{1}{l|}{Accuracy ($\mu$)} & $\sigma$  & \multicolumn{1}{l|}{Accuracy ($\mu$)} & $\sigma$  \\ \hline
\multirow{3}{*}{FEMNIST}            & Random                          & \multicolumn{1}{l|}{80.6}     & 0.1 & \multicolumn{1}{l|}{80.5}     & 0.2 & \multicolumn{1}{l|}{80.3}     & 0.2 & \multicolumn{1}{l|}{79.3}     & 0.5 & \multicolumn{1}{l|}{79.2}     & 0.3 \\ \cline{2-12} 
                                    & Ordered                         & \multicolumn{1}{l|}{80.6}     & 0.2 & \multicolumn{1}{l|}{80.5}     & 0.2 & \multicolumn{1}{l|}{80.4}     & 0.2 & \multicolumn{1}{l|}{80.3}     & 0.1 & \multicolumn{1}{l|}{79.7}     & 0.3 \\ \cline{2-12} 
                                    & \textbf{Invariant}                              & \multicolumn{1}{l|}{\color{red}\textbf{81.1}}     & 0.3 & \multicolumn{1}{l|}{\color{red}\textbf{80.9}}     & 0.1 & \multicolumn{1}{l|}{\color{red}\textbf{80.8}}     & 0.2 & \multicolumn{1}{l|}{\color{red}\textbf{80.3}}     & 0.4 & \multicolumn{1}{l|}{\color{red}\textbf{80.1}}     & 0.3 \\ \hline
\multirow{3}{*}{CIFAR10}            & Random                          & \multicolumn{1}{l|}{51.9}     & 1.3 & \multicolumn{1}{l|}{51.5}     & 0.5 & \multicolumn{1}{l|}{52.6}     & 0.3 & \multicolumn{1}{l|}{51.7}     & 1.0 & \multicolumn{1}{l|}{52.2}     & 0.3 \\ \cline{2-12} 
                                    & Ordered                         & \multicolumn{1}{l|}{52.3}     & 1.6 & \multicolumn{1}{l|}{51.6}     & 0.4 & \multicolumn{1}{l|}{54.6}     & 0.1 & \multicolumn{1}{l|}{53.1}     & 1.6 & \multicolumn{1}{l|}{53.0}     & 0.2 \\ \cline{2-12} 
                                    & \textbf{Invariant}                              & \multicolumn{1}{l|}{\color{red}\textbf{52.9}}     & 0.2 & \multicolumn{1}{l|}{\color{red}\textbf{53.0}}     & 0.1 & \multicolumn{1}{l|}{53.4}     & 0.2 & \multicolumn{1}{l|}{53.0}     & 0.2 & \multicolumn{1}{l|}{\color{red}\textbf{53.2}}     & 0.3 \\ \hline
\multirow{3}{*}{Shakespeare}        & Random                          & \multicolumn{1}{l|}{43.3}     & 0.1 & \multicolumn{1}{l|}{42.5}     & 0.1 & \multicolumn{1}{l|}{42.4}     & 0.1 & \multicolumn{1}{l|}{41.8}     & 0.2 & \multicolumn{1}{l|}{41.3}     & 0.1 \\ \cline{2-12} 
                                    & Ordered                         & \multicolumn{1}{l|}{42.9}     & 0.1 & \multicolumn{1}{l|}{42.3}     & 0.1 & \multicolumn{1}{l|}{42.2}     & 0.2 & \multicolumn{1}{l|}{41.9}     & 0.1 & \multicolumn{1}{l|}{41.4}     & 0.1 \\ \cline{2-12} 
                                    & \textbf{Invariant}                              & \multicolumn{1}{l|}{\color{red}\textbf{43.6}}     & 0.1 & \multicolumn{1}{l|}{\color{red}\textbf{42.5}}     & 0.1 & \multicolumn{1}{l|}{\color{red}\textbf{42.6}}     & 0.2 & \multicolumn{1}{l|}{\color{red}\textbf{42.2}}     & 0.2 & \multicolumn{1}{l|}{\color{red}\textbf{41.7}}     & 0.1 \\ \hline
\end{tabular}
}
\end{center}
\end{table*}

\section{Accuracy Evaluation}
We evaluated the accuracy of Invariant Dropout for different sub-model sizes and compared the results to the two baselines. 
Table~\ref{tab:accuracy} presents the average achieved accuracy ($\mu$) with standard deviation ($\sigma$) for three different datasets. 
%
%
Specifically, Invariant Dropout outperforms Random Dropout across all three datasets. 
As compared to Random Dropout, Invariant Dropout achieves a maximum accuracy gain of 1.5\% points and on average 0.7\% point higher accuracy for FEMNIST, 1.1\% point higher accuracy for CIFAR10 and 0.3\% point higher accuracy for Shakespeare datasets. 
Invariant Dropout also achieves a higher accuracy against Ordered Dropout across all three datasets.
As compared to Ordered Dropout, Invariant Dropout with a maximum of 1.4\% point increase in accuracy.
On average, Invariant dropout has 0.3\% point higher accuracy for FEMNIST, 0.2\% point higher accuracy for CIFAR10, and 0.4\% point higher accuracy for Shakespeare.
Invariant Dropout also achieves smaller $\sigma$  between runs of the same sub-model size and across all sub-model sizes.
Furthermore, accuracy values do not suddenly spike or drop significantly between sub-model sizes. 
The accuracy improvements of Invariant Dropout was verified to be significant at $r < 0.05$. 
%
%


\section{Computational Performance Evaluation}

\begin{figure}[t!]
    \centering
    \includegraphics[width=0.8\columnwidth]{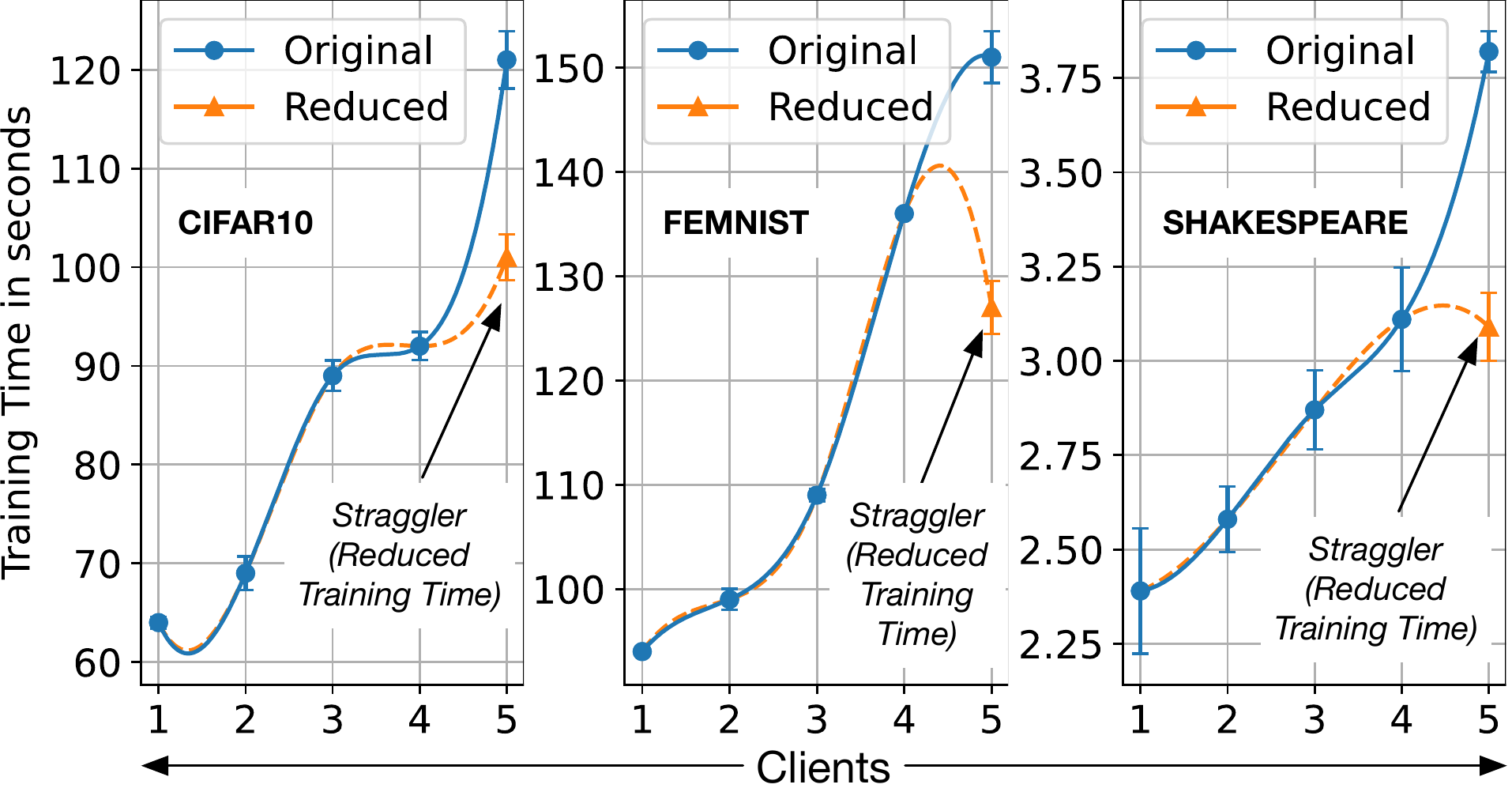}
   \caption[Straggler Training Time Reduction]{The reduction in training time for stragglers before and after applying Invariant Dropout. Without Invariant Dropout, the straggler would execute between 10\% to 32\% of the target time. After applying Invariant Dropout, the straggler runs within 10\% of the target time.}
    \label{fig:executiontime}
\end{figure}

We evaluated the training time reduction when Invariant Dropout dynamically selects the sub-model for the straggler. 
Figure~\ref{fig:executiontime} shows that Invariant Dropout correctly selects the sub-model size so that the stragglers' training time is almost similar to that of the next slowest client. 
Without Invariant Dropout, the straggler would execute between 10\% to 32\% of the target time.
After applying Invariant Dropout, the straggler runs within 10\% of the target time. 

In general, the accuracy results show that in general, when the straggler trains with a larger sub-model size, the overall accuracy for the global model is higher compared to that of a smaller sub-model size. 
Thus, Invariant Dropout chooses the largest possible sub-model size that minimizes training time variance across clients.  

It should be noted that, executing the Invariant Dropout algorithm also does not add significant processing overhead to the overall training time. 
This is because, the additional computation required such as creating and transforming sub-models for aggregation, are all done on a centralized server instead of edge devices. 
We empirically determined that this overhead on the server is negligible compared to the training reduction -- 50ms versus 0.8 seconds to 26 seconds. 

\section{Scalability Studies}

\begin{figure*}[t!]
    \centering
    \includegraphics[width=1\columnwidth]{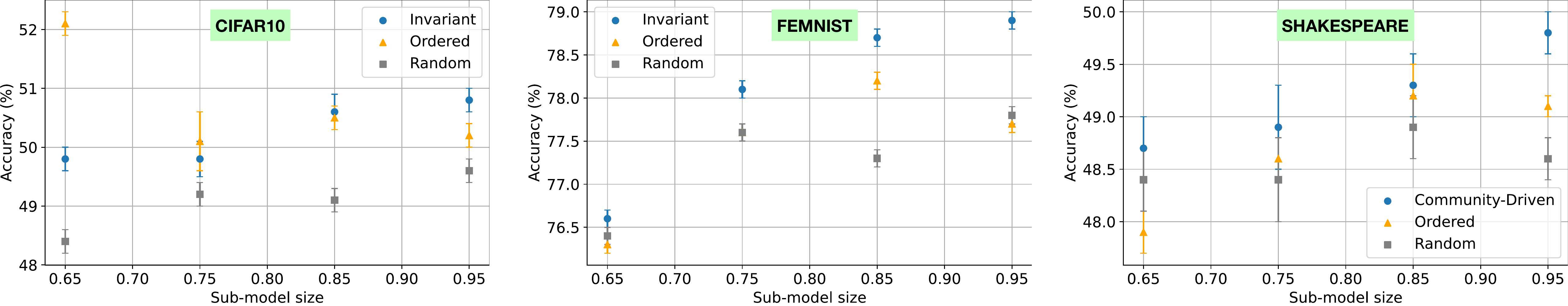}
   \caption[Invariant Dropout Accuracy in Scale]{The accuracy comparison of Invariant Dropout with Ordered and Random Dropout as we scale to 50-100 clients with 20\% of the slowest clients being stragglers. This study simulates a large number of clients on a server cluster. Overall, Invariant Dropout consistently outperforms Ordered and Random Dropout and maintains a better accuracy profile.}
    \label{fig:scale}
\end{figure*}

In order to evaluate the scalability of Invariant Dropout, we tested the technique on a large number of simulated clients (50 to 100), also implemented using the Flower Framework.
For the scalability tests, we run all clients on NVIDIA Tesla V100 GPUS clusters, each machine running 10 to 20 clients in parallel. 
We identify 20\% of the slowest clients as stragglers.
Figure~\ref{fig:scale} shows the accuracy performance across all three datasets. 
Overall, Invariant Dropout consistently outperforms Ordered and Random Dropout and maintains a better accuracy profile similar to Table~\ref{tab:accuracy}.

\chapter{Related Work}
\label{ch:Related}
In the domain of heterogeneous client optimization, several prior work have tried to mitigate the effects of stragglers.

\section{Server Offloading Strategies} 
To accelerate local training on computationally weak devices, Wu et. al., Ullah et. al., and Wang et. al. propose offloading layers of deep neural networks to servers~\cite{FedAdapt, FedFly, FedLite}.

Wu et. al.~\cite{FedAdapt} focused on reducing training time with heterogeneous networks with an adaptive mechanism that offloads a part of the model to the server using split learning. Additionally, the technique also adaptive selects the layers of the model that should be offloaded to the server based on straggler capabilities. 

Ullah et. al.~\cite{FedFly} focused on device mobility while performing Federated learning and offloading training to edge servers. When a edge device participates in Federated Learning, it is possible that the client might move from one location to another. This would cause the client to move out of range from the current edge server -- this needs to be part of the offloading decision.

Wang et. al.~\cite{FedLite} introduces a compression method for split learning (offloading to the server). This aims to reduce the communication overheads for computation and communication resource-limited edge clients.

Broadly, Invariant Dropout can be easily applied on the top of these techniques to determine which part of the model needs to be offloaded to the server.

\section{Communication Optimizations}
To minimize overall learning time, Han et. al. have proposed an online learning algorithm that determines the degree of sparsity based on communication and computation capabilities~\cite{online}. 

PruneFL~\cite{PruneFL} introduces an approach to adaptive select model sizes during FL to reduce communication and computation overhead and overall training time. 

Chen et. al. propose a communication-efficient FL framework, using a probabilistic device selection method, to select clients that benefit the convergence speed and training loss~\cite{PNAS}. As well, the authors also proposed a quantization method to reduce communication overheads.

Xu et. al. propose a framework that incorporates overhead reduction techniques to enable efficient model training on resource limited edge devices~\cite{FLPQSU}. The algorithm performs optimization techniques like pruning, quantization, and prevents less helpful updates to the server. 

Compared to these prior work, Invariant Dropout presents a new insight that avoids communication and computation overheads to `invariant' neurons.


\chapter{Conclusions and Future Work}
\label{ch:Conclusions}

Rapid technology development and variation among handheld devices has resulted in some devices becoming stragglers. Straggler devices offer low computational performance and act as bottlenecks in Federated Learning. They not only increase the training time but also reduce the accuracy of the ML model. This thesis mitigates the performance and accuracy effects of stragglers by dynamically creating tailored sub-models that contain only the neurons that change above a certain threshold. Our technique, called Invariant Dropout, mitigates performance overheads while providing a 1.4\% point higher accuracy than the state-of-the-art technique, called Ordered Dropout, across three different datasets and ML models.

\section{Future Work}
Based on this work, we identified two potential areas of further study
\subsection{Extensions of Invariant Dropout}
This work has implemented Invariant Dropout with three datasets and experimented with five mobile clients. In future works based on Invariant Dropout, we could investigate how the technique could be extended to support a wider variety of datasets and models. As well, another potential area of study is to explore the best way to set performance target time when there are multiple stragglers in the FL network. It is also possible to customize the target time and sub-model size for each identified straggler client. 

\subsection{Drop-Threshold Identification Improvements}
Currently, the Invariant Dropout technique initializes the drop-threshold based on characteristics of the early training iterations, and simply increments the threshold by a fixed value until the number of invariant neurons is satisfied. While this method is simple and requires minimum computing overhead, depending on the values of the initial threshold and the fixed increment, the threshold might settle at a value too large for larger sub-model sizes, or reach the suitable threshold too slowly for smaller sub-model sizes. Hence the threshold setting algorithm could be further improved to search for the target threshold in an even more dynamic fashion. 


\begin{singlespace}
\raggedright
\bibliographystyle{abbrvnat}
\bibliography{ref}
\end{singlespace}

\backmatter

\end{document}